\documentclass{article}

\usepackage{PRIMEarxiv}

\usepackage{pmboxdraw}
\usepackage[utf8]{inputenc} 
\usepackage[T1]{fontenc}    
\usepackage{hyperref}       
\usepackage{url}            
\usepackage{booktabs}       
\usepackage{amsfonts}       
\usepackage{nicefrac}       
\usepackage{microtype}      
\usepackage{lipsum}
\usepackage{fancyhdr}       
\usepackage{graphicx}       
\graphicspath{{media/}}     
\usepackage{subcaption}
\usepackage{tabularx}
\usepackage{float}

\usepackage{hyperref}   
\usepackage{url}        
\usepackage{doi}        

\usepackage{colortbl}
\usepackage{xcolor}
\usepackage{adjustbox}

\definecolor{lightgreen}{rgb}{0.8,1,0.8}
\definecolor{lightblue}{rgb}{0.67, 0.9, 1}
\definecolor{lightorange}{rgb}{1, 0.8, 0.67}
\definecolor{lightyellow}{rgb}{1, 1, 0.8}
\definecolor{lightred}{rgb}{1, 0.5, 0.5}

\title{ Qtok: A Comprehensive Framework for Evaluating Multilingual Tokenizer Quality in Large Language Models
}

\author{
  Iaroslav Chelombitko, Egor Safronov, Aleksey Komissarov \\
  Neapolis University Pafos \\
  Paphos, Cyprus\\
  \texttt{ad3002@gmail.com} \\
}

\begin{document}
\maketitle

\begin{abstract}
In the development of Large Language Models (LLMs), considerable attention has been given to the quality of training datasets. However, the role of tokenizers in the LLM training pipeline, particularly for multilingual models, has received less focus. The quality of tokenization can significantly impact a model's ability to handle diverse languages effectively. We introduce Qtok, a tool designed to assess tokenizer quality with a specific emphasis on their performance in multilingual contexts.

Our research proposes a set of metrics for evaluating tokenizer quality, including measures of language coverage, token completeness, and distribution across languages and linguistic categories. Qtok applies these metrics to evaluate 13 distinct tokenizers from 58 publicly available models, analyzing their output across different linguistic contexts. Our analysis revealed significant variations in token distribution across languages and categories, highlighting potential biases and areas for improvement in current tokenization strategies.

This research contributes to the field of tokenizer evaluation within multilingual LLM development by providing a systematic approach to assessing tokenizer quality. Our findings highlight the critical role of tokenization in multilingual LLM capability. The Qtok tool and our analysis methodology offer practical means for researchers to evaluate and improve tokenization strategies for multilingual applications. We offer a method to compare tokenizer quality across these metrics, which may be useful when selecting or adjusting tokenizers for specific multilingual LLM applications.

Qtok availability: https://github.com/nup-csai/Qtok/

\end{abstract}


\section{Introduction}

In recent years, Large Language Models (LLMs) have demonstrated remarkable capabilities across a wide range of natural language processing tasks. While much attention has been given to factors such as dataset quality (Chen \& Mueller, 2024), model architectures, and training objectives (Raiaan et al., 2024), the crucial role of tokenizers in LLM development, particularly for multilingual models, has been relatively overlooked. Tokenization serves as the fundamental bridge between human-readable text and the numerical representations that machine learning models can interpret, significantly impacting model performance, training efficiency, and language handling abilities (Ali et al., 2024; Schmidt et al., 2024).

Recent studies have highlighted the importance of tokenizer choice in LLM training. (Ali et al., 2024) demonstrated that tokenizer selection can significantly affect both model performance and training costs, especially in multilingual contexts. (Rajaraman et al., 2024) provided theoretical insights into the role of tokenization in LLMs, showing how appropriate tokenization enables even simple models to effectively capture complex language distributions (Limisiewicz et al., 2023).

Our research builds on these ideas and offers the following contributions to how we assess and improve tokenizers for multilingual LLMs (Figure 1):

\begin{enumerate}
    \item \textbf{Token Classification Framework:} We join various ideas about token classification from existing literature and propose a comprehensive framework for categorizing tokens. This classification system provides a foundation for more nuanced analysis of tokenizer performance across different linguistic contexts.
    
    \item We introduce \textbf{Qtok}, a tool designed to evaluate tokenizer quality, with a focus on multilingual performance. Qtok offers metrics that assess tokenizers both at the system level and for individual tokens, examining factors like token completeness, accuracy, and distribution across different languages and categories. These metrics go beyond traditional measures, such as character-to-token efficiency, to provide insights that link tokenizer properties with downstream model performance.
    
    \item One of the key contributions of our work is the introduction of the concept of \textbf{core tokens}. These are tokens that consistently appear across different tokenizers, grouped by the vocabulary size, regardless of the specific model or dataset from which they are derived. This approach provides a foundation for a more streamlined and standardized tokenization process, which can enhance compatibility and reduce token variability between different models.
    
    \item \textbf{Comprehensive Tokenizer Evaluation:} With Qtok, we performed an in-depth evaluation of popular tokenizers from the Chatbot Arena dataset (Chiang et al., 2024), identifying 13 distinct tokenizers across 58 models. This analysis highlights the strengths and weaknesses of various tokenization methods in different linguistic contexts, with a particular emphasis on non-English language representation.
\end{enumerate}

\begin{figure}[htbp]
    \centering
    \includegraphics[width=1\linewidth]{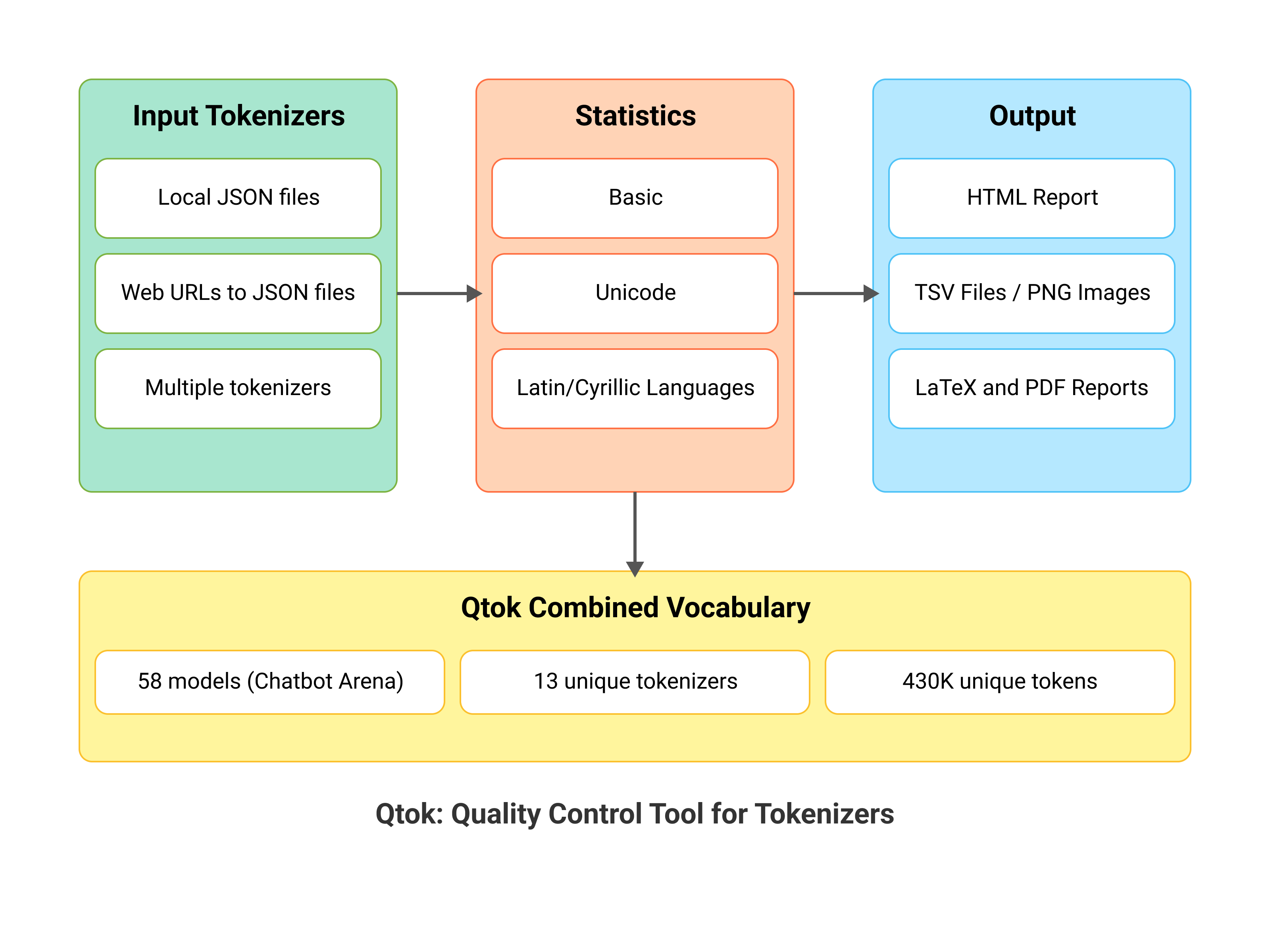}
    \caption{Graphical abstract of the proposed Qtok evaluation tool.}
    \label{fig:fig1}
\end{figure}

\section{Results}
\label{sec:headings}

\subsection{There are 13 distinct tokenizers in 58 models from Chatbot Arena}

As of September 2024, licensing restrictions severely limited the available pool of tokenizers. Within the Chatbot Arena dataset, a significant number of models operate under closed licenses, restricting access to their tokenizers and limiting their inclusion in this study. This licensing issue substantially reduced the number of tokenizers available for analysis (Table 1).

Furthermore, even among models with open licenses, the lack of official repositories on platforms such as Hugging Face posed additional obstacles. To maintain the integrity of our analysis, we made the methodological decision to use only tokenizers from official repositories, eschewing potentially altered user-created versions. This criterion further narrowed our dataset but ensured the authenticity of the tokenizers under examination.
Format incompatibility emerged as another significant challenge. Several models did not provide tokenizers in the standard Hugging Face format, which typically requires the config.json, tokenizer.json, and tokenizer\_config.json files. This incompatibility complicated the retrieval and further analysis of these tokenizers.

After applying these stringent criteria to the models listed on Chatbot Arena (formerly LMSYS) we selected only those with officially accessible tokenizers that met our requirements for format compatibility and ease of access. This rigorous selection process resulted in a final dataset of 58 models, which formed the basis of our further analysis.

LLM models can be categorized into various types, such as general-purpose LLMs, code-focused LLMs, and others. For this study, we limited our analysis to models designed for processing natural language, thus excluding those specialized for code-related tasks, as they necessitate a different methodological approach. 

Additionally, the absence of certain tokenizers from the Transformers library (Wolf et al., 2020) required individual exploration and, in some cases, reverse-engineering to gain access. This process was both time-consuming and technically challenging, ultimately limiting the scope of our analysis. As a result, we excluded these restricted models from our initial evaluation.

\begin{table}[h]
\centering
\caption{Model classification according to its openness and standardized accession to weights and tokenizer (September 2024).}
\begin{tabular}{lcc}
\toprule
\textbf{License} & \textbf{Number of models} & \textbf{Number of open models} \\
\midrule
AI2 ImpACT Low-risk & 1 & 0 \\
Apache 2.0 & 20 & 20 \\
CC-BY-NC-4.0 & 7 & 7 \\
CC-BY-NC-SA-4.0 & 3 & 3 \\
DBRX LICENSE & 1 & 1 \\
DeepSeek & 2 & 2 \\
DeepSeek License & 2 & 2 \\
Falcon-180B TII License & 1 & 1 \\
Gemma license & 7 & 7 \\
Jamba Open & 2 & 2 \\
Llama 2 Community & 10 & 10 \\
Llama 3 Community & 2 & 2 \\
Llama 3.1 Community & 4 & 4 \\
Llama 3.2 & 2 & 2 \\
MIT & 9 & 9 \\
Mistral Research & 1 & 1 \\
NVIDIA Open Model & 1 & 1 \\
Non-commercial & 7 & 7 \\
Other & 1 & 1 \\
Proprietary & 52 & 0 \\
Qianwen LICENSE & 8 & 8 \\
Qwen & 1 & 1 \\
Yi License & 1 & 1 \\
\bottomrule
\end{tabular}
\end{table}

Among these 58 models, there are models that belong to the same family. This means that models within the same family will share an identical tokenizer. Additionally, the models we are analyzing include those that are fine-tuned versions of other models. In this case, the tokenizers are almost identical, differing only by the presence of additional reserve (control) tokens. In this study, such tokenizers are considered identical because the vocabulary tokens remain unchanged.

\begin{figure}[H]
    \centering
    \includegraphics[width=1\linewidth]{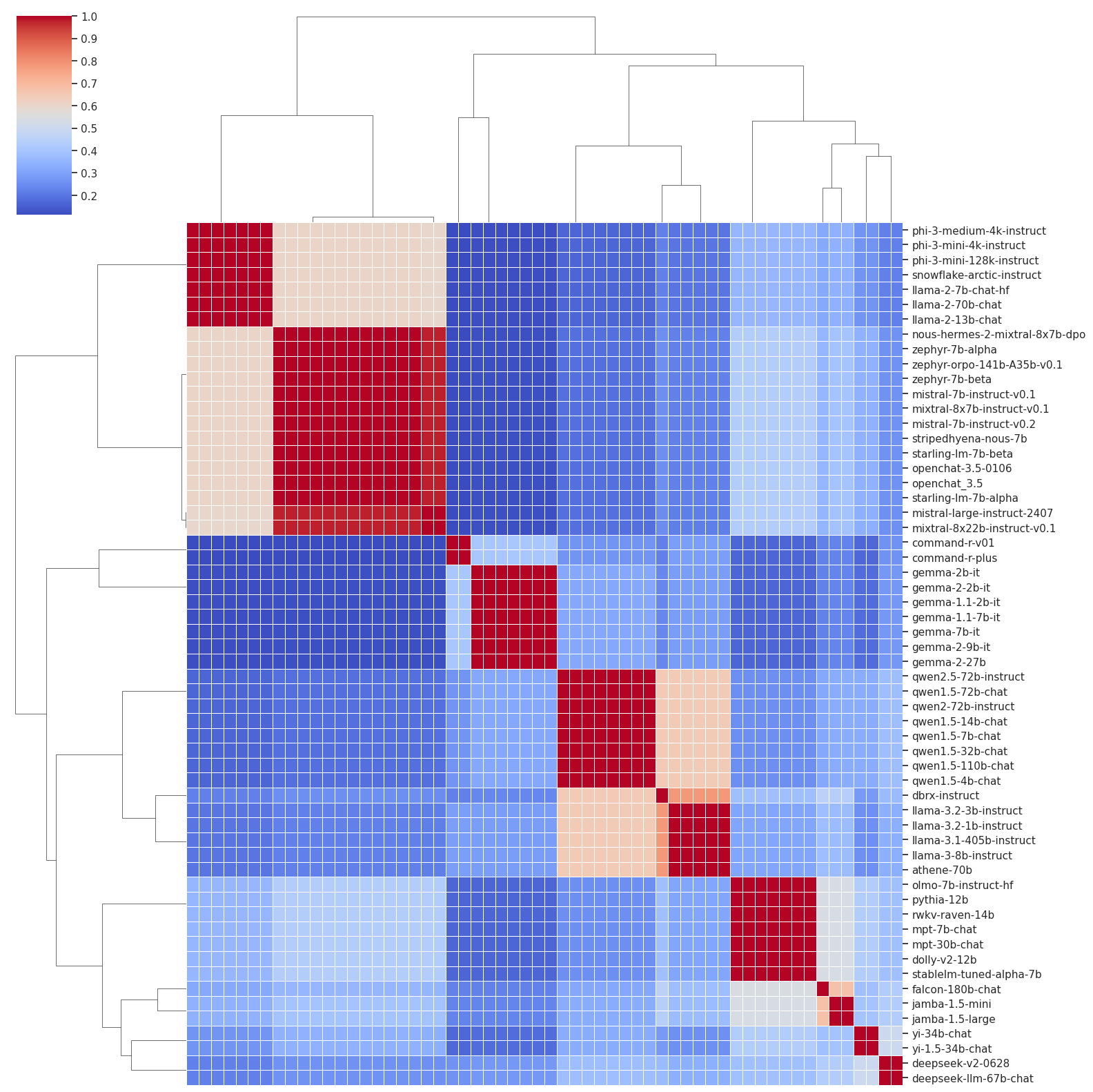}
    \caption{Weighted Jaccard Similarity between vocabularies in different models.}
    \label{fig:enter-label1}
\end{figure}

Before clusterization of tokenizers, we normalized tokens that indicate the start of a word. In our process of normalizing spaces before tokens, we addressed inconsistencies in token representations across various tokenizers. We identified three special markers for word starts: `\pmboxdrawuni{2581}`, `Ġ`, and `\texttt{\textbackslash u0120}`, which represent space prefixes in different tokenization systems. These markers were unified by replacing them with `Ġ` to maintain consistency across tokenizers.

After normalization, we identified tokens in the vocabulary that start with space indicators (`Ġ`) and ensured that decoded tokens also begin with a space. This normalization guarantees that all space-prefixed tokens are handled uniformly when using the tokenizer for subsequent tasks, thereby preventing discrepancies in token boundaries.

We discovered that there are two methods to encode raw bytes in tokenizers: one using codes like <0xff>, and another following the mapping suggested by OpenAI, which was later adopted by the Hugging Face Tokenizer library written in Rust (Moi \& Patry, 2023). To handle this, we implemented custom code for decoding tokens encoded by the Hugging Face Tokenizer library. This allowed us to normalize tokens to a consistent representation in UTF-8 across all tokenizers. This normalization ensures that the tokenization process remains uniform, regardless of the tokenizer used.

To cluster tokenizers we used the weighted jaccard similarity (Figure 2). We found several groups of models according to the dendrogram that can be explained by model generation and by vocabulary size. In the top left quadrant, the cluster includes models from llama-2 and mixtral family. Llama-2 related models includes `snowflake-arctic-instruct` (Merrick et al., 2024), `llama-2-7b-chat`, `llama-2-70b-chat` (Touvron, Lavril, et al., 2023, p. 2; Touvron, Martin, et al., 2023, p. 2),  `phi-3-mini-4k-instruct`, `phi-3-mini-128k-instruct`, `phi-3-medium-4k-instruct` (Abdin et al., 2024). Mixtral related models includes `mixtral-8x22b-instruct-v0.1` (Mistral AI team, 2024a), `zephyr-7b-beta` (Tunstall et al., 2023), `mixtral-8x7b-instruct-v0.1` (Jiang et al., 2024), `openchat-3.5-0106` (Wang et al., 2024), `starling-lm-7b-beta` (Zhu et al., 2023), `mistral-7b-instruct-v0.2` (Jiang et al., 2023), and others, Both llama-2 and mixtral can be called the second generation of models.

In the middle of the heatmap is located the third generation of the models, the Gemma and Command models that form another distinct cluster. Models such as `gemma-2-7b`, `gemma-7b-it` (Gemma Team et al., 2024), and `command-r-v01` (Command R: RAG at Production Scale, 2024) display a high degree of similarity. According to dendrogram Qwen (Bai et al., 2023) and Llama-3 (AI@Meta, 2024, p. 3) models also can be placed in the group.  

Separate group of models located towards the lower part of the heatmap, a group of models including pythia (Biderman et al., 2023), olmo (Groeneveld et al., 2024), mpt (Introducing MPT-7B, 2023; MosaicML NLP Team, 2023) and others, they can originally be traced back to the first generation of llama models.

Other clusters formed include the `dolly-v2-12b` (Conover et al., 2023) , `deepseek-lm-67b-chat` (DeepSeek-AI et al., 2024), `falcon-180b-chat` (Almazrouei et al., 2023), and various `jamba` (Lieber et al., 2024) models.

Additionally, there are a few models that are more isolated, appearing in distinct parts of the heatmap. Models such as `athene-70b` (Nexusflow, 2024), `mpt-30b-chat`, and `mistral-large-instruct-2407` (Mistral AI team, 2024b) are less closely related to the larger clusters.

We use a simplified definition of identity between tokenizers. This approach gives us 13 distinct tokenizers from 58 models. The full list of 13 tokenizers is shown in Table 2.

\begin{table}[h]
\centering
\caption{Tokenizers groups.}
\begin{tabular}{cccccc}
\toprule
\textbf{ID} & \textbf{Group} & \textbf{Group Size} & \textbf{Tokens} & \textbf{Release Year} \\
\midrule
1  & llama-2    & 7  & 31999   & 2023 \\
2  & mistral    & 14 & 32000   & 2023 \\
3  & mixtral    & 2  & 32768   & 2024 \\
4  & pythia     & 7  & 49790   & 2023 \\
5  & yi-34b     & 2  & 63985   & 2023 \\
6  & falcon     & 1  & 64456   & 2023 \\
7  & jamba      & 2  & 65536   & 2024 \\
8  & deepseek   & 2  & 99467   & 2024 \\
9  & dbrx       & 1  & 99579   & 2024 \\
10 & llama-3    & 5  & 126784  & 2024 \\
11 & qwen       & 8  & 150307  & 2023 \\
12 & command-r  & 2  & 253266  & 2024 \\
13 & gemma      & 7  & 255999  & 2024 \\
\bottomrule
\end{tabular}
\end{table}

The dendrogram and heat map according to weighted jaccard similarity between 13 tokenizers is evident that the tokenizers grouped into distinct clusters. The fact that Llama 2 merges into the same group as Mistral and Mixtral, that the more recent models, such as Command-R and Gemma, also grouped together, that Llama 3 is in the same group as Qwen, and that the Pythia model actually belongs to the first generation of Llama models, clearly shows that different generation of models have different tokenizers with the trend of tokenizer enlarging (Figure 3).

\begin{figure}[H]
    \centering
    \includegraphics[width=1\linewidth]{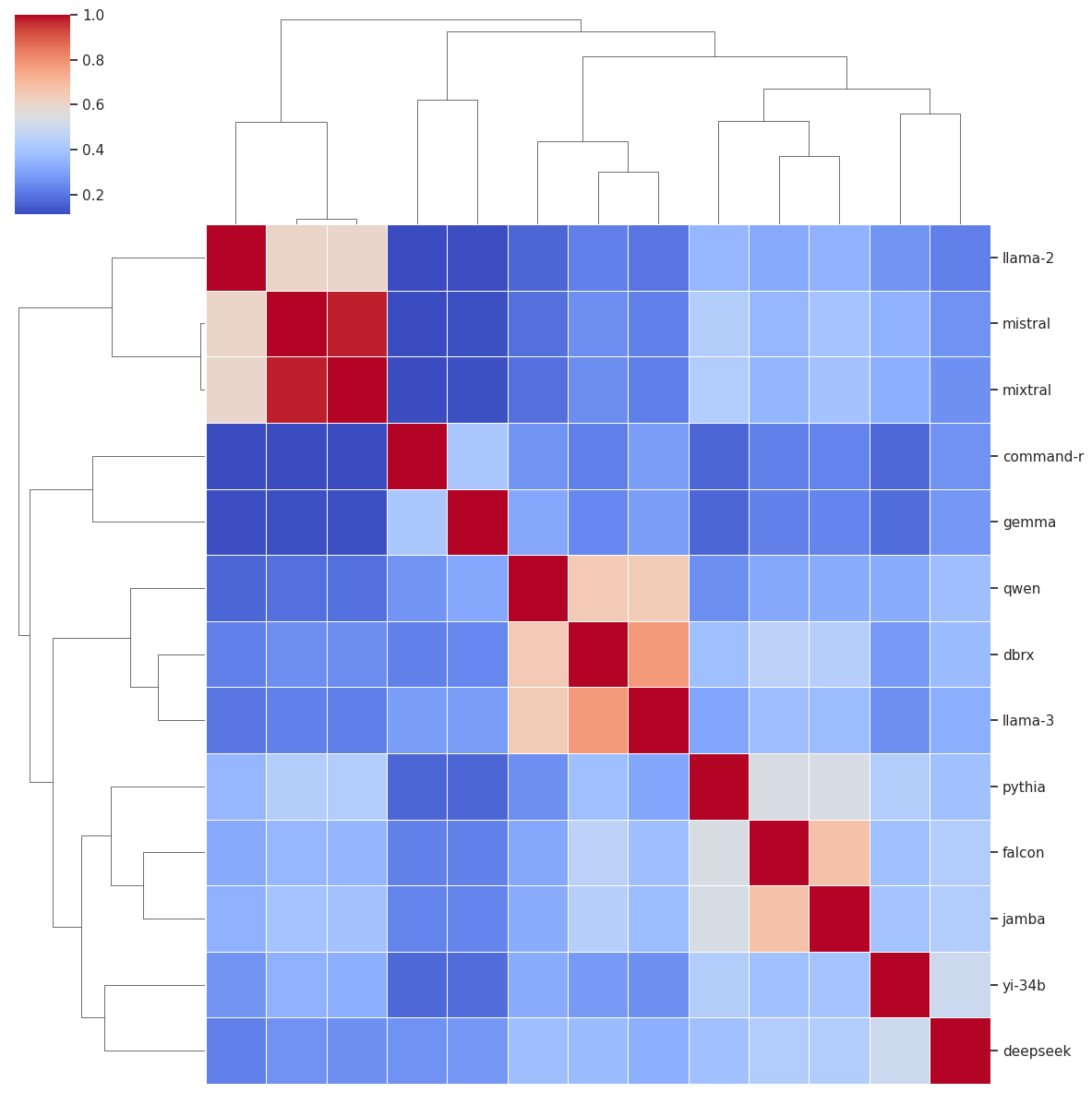}
    \caption{Weighted Jaccard Similarity Between Tokenizer Vocabularies. A comprehensive heatmap and accompanying dendrogram depicting the weighted Jaccard similarity between the vocabularies of thirteen distinct tokenizers. Note that mixtral and mistral have different tokenizers despite their high correlation coefficient of 0.9.}
    \label{fig:enter-label2}
\end{figure}

\subsection{Unified tokenizer}

To avoid analyzing each tokenizer individually, we decided to combine all the tokens into a unified tokenizer. This artificial unified tokenizer allows us to examine and annotate tokens independently of any specific tokenizer. Later, this unified tokenizer can be used as a metric for more detailed analysis of individual tokenizers.

The process of consolidating tokens into a single unified tokenizer is relatively straightforward. First, the vocabulary from each unique tokenizer is collected. Then, for each token in an individual tokenizer’s vocabulary, a check is performed to determine whether it is already present in the collective set of tokens.  If the token is absent, it is added to the unified file; if it is already included, it is disregarded. Additionally, for each token, we maintain a list of its ranks across all 13 tokenizers. This allows us to track the relative importance and usage of each token within different tokenization systems. For individual models within the same group, there may be slight differences in tokens, primarily due to instruction-tuning, which introduces additional tags. However, we chose not to account for these specific tokens, and as a result, the total number of tokens across tokenizers is approximate rather than absolute.

This approach ensures that the "unified tokenizer" consists exclusively of non-duplicate (i.e., unique) tokens, enabling more effective general analysis. Moreover, only unique tokenizers were used to construct this overall vocabulary. We had a question regarding how much tokenizers contribute to the overall addition of tokens. To explore this, we analyzed the cumulative growth of our unified tokenizer as we added new tokenizers to it. The tokenizers were sorted by size, allowing us to observe how each new tokenizer impacts the total number of unique tokens. We found the rapid advancement in language model development, with newer models incorporating significantly larger token vocabularies. This trend likely correlates with improvements in model performance, capability, and potentially, the ability to understand and generate more nuanced and diverse language. The exponential growth also raises questions about the computational resources required for training and running these increasingly large models, as well as their potential capabilities and limitations (Figure 4).

\begin{figure}[H]
    \centering
    \includegraphics[width=1\linewidth]{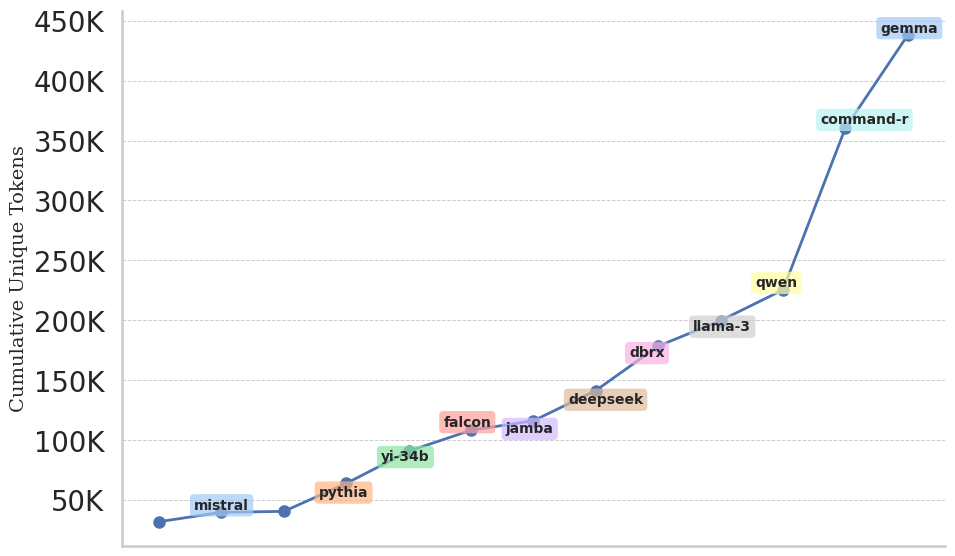}
    \caption{Cumulative Number of Unique Tokens by Model.}
    \label{fig:enter-label3}
\end{figure}

Figure 4 illustrates the cumulative growth of unique tokens across language models. Initially, models like Mistral and Pythia show a plateau (30,000-50,000 tokens), suggesting a focus on core vocabulary. Mid-range models (Yi-34b, Falcon, Jamba, Deepseek) exhibit steeper growth, reaching ~150,000 tokens, indicating expanded language coverage.
A marked acceleration occurs from the dbrx model onward. Llama-3, Qwen, Command-R, and Gemma show substantial increases, with Gemma peaking at ~440,000 unique tokens. This trend suggests a shift towards more comprehensive language representation, potentially enabling better handling of diverse linguistic phenomena but also increasing computational demands.

\subsection{Token Annotation and Classification Without Language Association}

With the unified tokenizer in place, we were able to begin the process of token classification. This allowed us to determine how to group the ~430,000 tokens into key categories. 

Firstly we had an attempt to implement filtering using a lemmatizer. However, we encountered the issue that lemmatizer lists are available for only a few languages, while the tokenizer supports many more. Consequently, standard lemmatizers like those from spaCy (Matthew Honnibal et al., 2020) or NLTK (Bird et al., 2009) do not suffice. As a result, we abandoned this approach and decided to implement a more sophisticated nonlanguage-specific processing method, which we will detail below.

Token classification starts with closed lists, such as control tags, since these are easily identifiable and can be excluded without much effort. Next, we handle bytes encoded as four-byte characters (from 0 to 255) as a sanity check, ensuring that the tokenizer can theoretically cover any input. After that, we process single characters, both ASCII and non-ASCII, as their number is finite and manageable. Following this, we focus on tokens containing flanking bytes that do not form valid characters.

Once these stages are complete, we filter out code-related elements, such as equal signs, bracket sequences, and carriage returns—anything typically associated with code. What remains are tokens that hold semantic meaning, which are divided into three categories:

\begin{enumerate}
    \item Tokens that \textbf{start with a space}, typically representing the beginning or complete words.
    \item Tokens that \textbf{do not start with a space}, usually fragments of words, such as their middle or end.
    \item Individual character tokens.
\end{enumerate}

We applied this classification process across 13 tokenizers and identified the main token groups, providing a structured foundation for further analysis (Table 3).

\begin{figure}[H]
    \centering
    \includegraphics[width=1\linewidth]{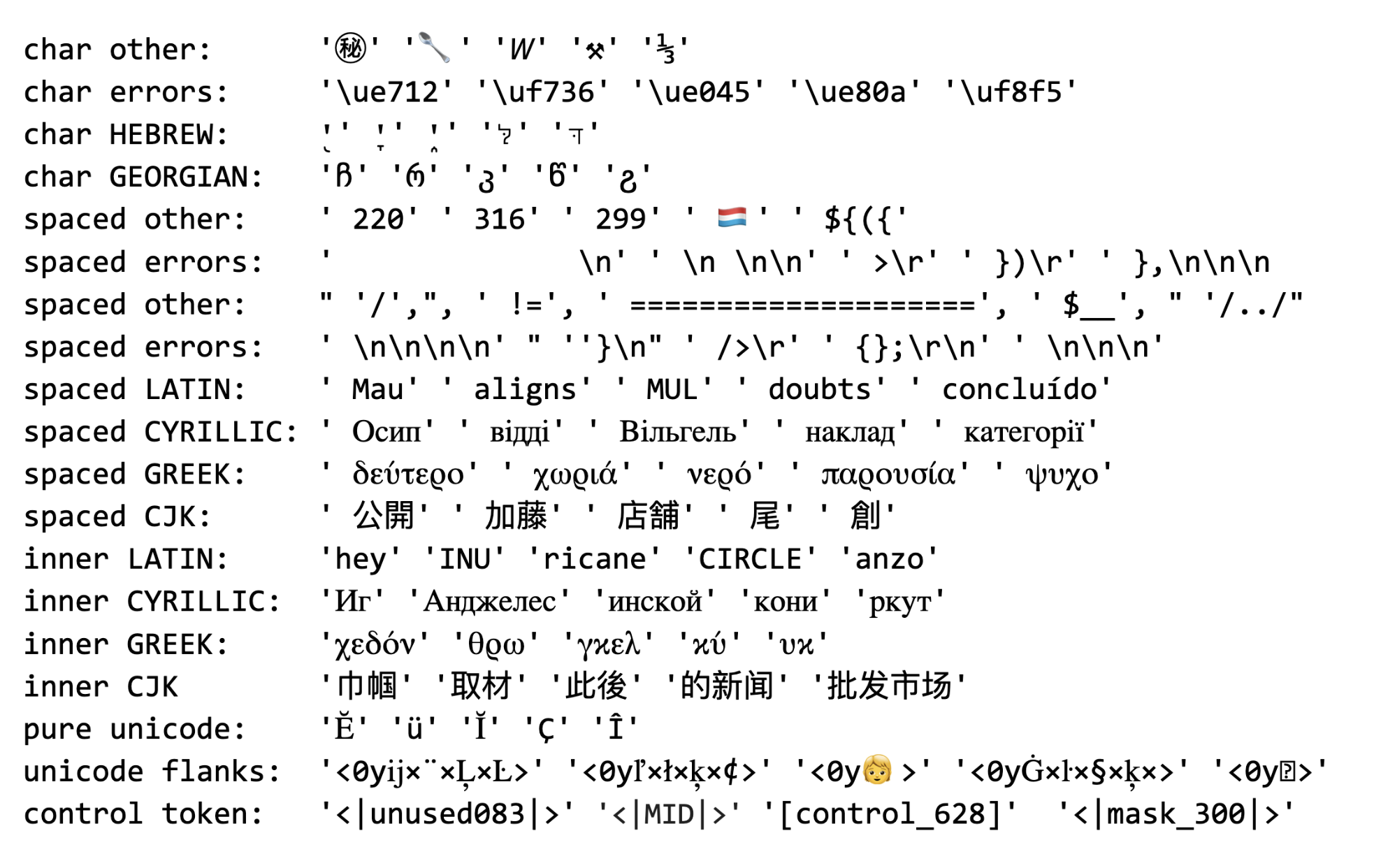}
    \caption{Token Examples by Category.}
    \label{fig:enter-label4}
\end{figure}

\begin{table}[h]
    \centering
    \caption{Token classification in 13 tokenizers (in percents).}
    \begin{tabularx}{\textwidth}{lXXXXXXXXXXXXX}
        \toprule
        \textbf{Tokenizer} & \textbf{control tokens} & \textbf{pure unicode} & \textbf{char alpha} & \textbf{spaced alpha} & \textbf{inner alpha} & \textbf{char other} & \textbf{spaced other} & \textbf{inner other} & \textbf{unicode flanks} & \textbf{char errors} & \textbf{spaced errors} & \textbf{inner errors} \\
        \midrule
        Qtok        & 0.68 & 0.06 & 4.38 & 44.97 & 38.55 & 0.93 & 0.83 & 7.05 & 1.35 & 0.21 & 0.19 & 0.81 \\
        llama-2     & 0.05 & 0.80 & 5.09 & 50.15 & 38.66 & 0.80 & 1.09 & 3.25 & 0    & 0.04 & 0.02 & 0.06 \\
        mistral     & 0.02 & 0.80 & 8.47 & 48.20 & 37.09 & 0.91 & 1.02 & 3.13 & 0    & 0.21 & 0.03 & 0.12 \\
        mixtral     & 2.36 & 0.78 & 8.27 & 47.07 & 36.22 & 0.89 & 0.99 & 3.06 & 0    & 0.20 & 0.03 & 0.12 \\
        pythia      & 0.01 & 0.37 & 1.37 & 54.27 & 33.81 & 0.13 & 2.91 & 5.57 & 1.02 & 0.13 & 0.07 & 0.35 \\
        yi-34b      & 0.30 & 0.40 & 6.57 & 37.35 & 52.43 & 0.17 & 0.54 & 2.21 & 0    & 0.02 & 0    & 0.01 \\
        falcon      & 0.01 & 0.26 & 1.84 & 61.26 & 31.65 & 0.09 & 0.08 & 3.14 & 1.45 & 0.03 & 0.03 & 0.16 \\
        jamba       & 2.36 & 0.39 & 3.13 & 59.50 & 31.60 & 0.45 & 0.61 & 1.79 & 0    & 0.06 & 0.02 & 0.07 \\
        deepseek    & 0.03 & 0.18 & 3.66 & 46.95 & 45.05 & 0.11 & 0.71 & 2.39 & 0.56 & 0.04 & 0.01 & 0.32 \\
        dbrx        & 0.03 & 0.18 & 1.04 & 42.48 & 28.55 & 0.10 & 1.19 & 22.14 & 0.78 & 0.03 & 0.64 & 2.84 \\
        llama-3     & 0.21 & 0.16 & 2.80 & 43.23 & 30.47 & 0.16 & 1.02 & 18.04 & 1.06 & 0.05 & 0.51 & 2.27 \\
        qwen        & 0.01 & 0.17 & 9.63 & 33.62 & 36.21 & 1.92 & 0.79 & 14.37 & 0.96 & 0.02 & 0.42 & 1.88 \\
        command-r   & 0.02 & 0.08 & 2.31 & 57.24 & 36.14 & 0.07 & 0.44 & 1.78 & 1.76 & 0.01 & 0.04 & 0.11 \\
        gemma       & 0.06 & 0.10 & 6.14 & 48.23 & 40.19 & 1.50 & 0.66 & 2.62 & 0    & 0.36 & 0.03 & 0.11 \\
        \bottomrule
    \end{tabularx}
\end{table}

As a result, we produce the twelve groups of tokens (in order of applying filters) (see token examples in Figure 5 and basic statistics in Figure 6),

\begin{figure}[H]
    \centering
    \includegraphics[width=1.0\linewidth]{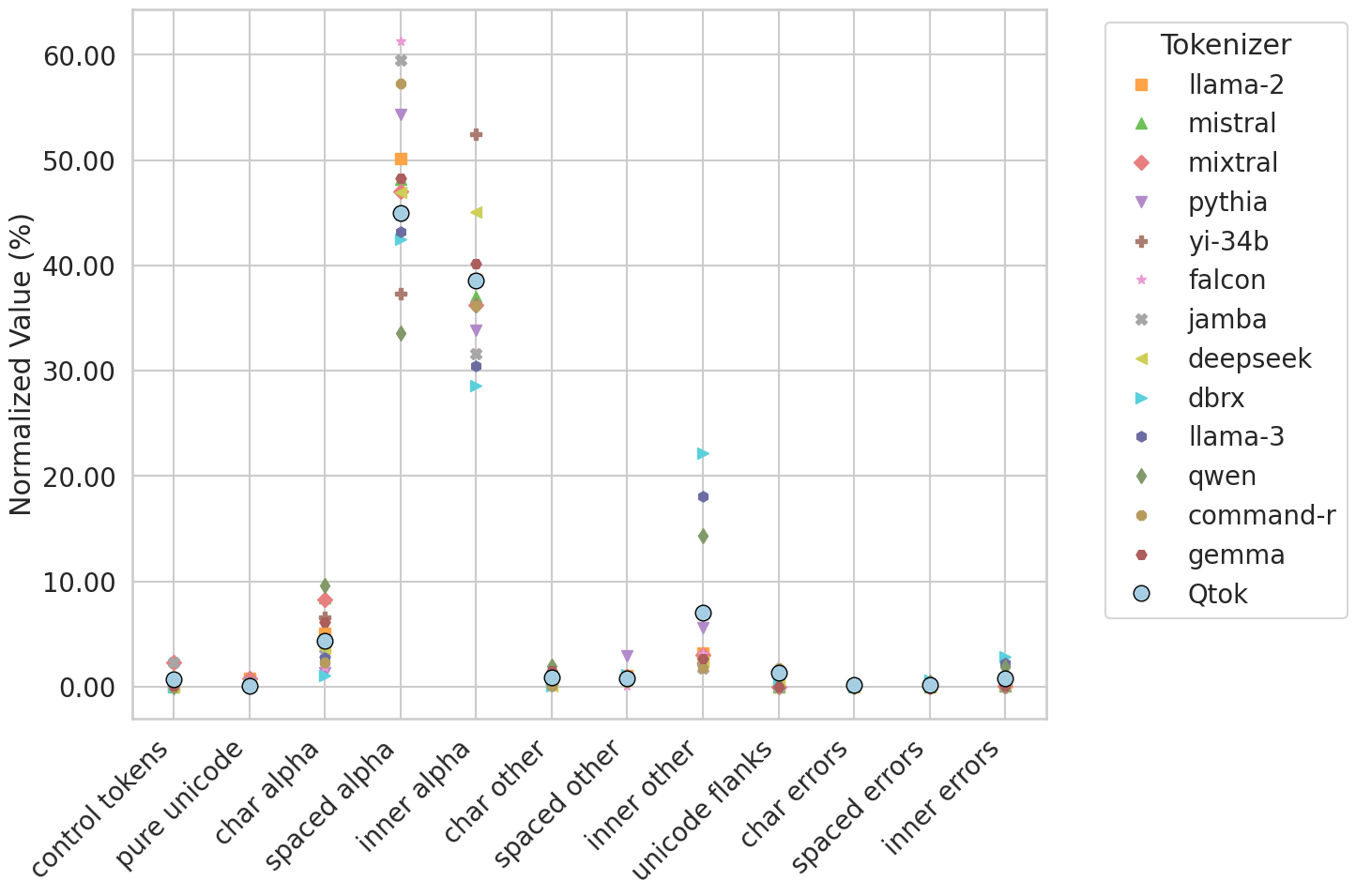}
    \caption{Statistics for token classes}
    \label{fig:enter-label5}
\end{figure}

Our analysis of 13 distinct tokenizers, including the unified Qtok tokenizer, revealed significant variations in token distribution across different categories (Figure 6). This classification provides insights into how various tokenizers represent language elements, with particular implications for multilingual and ideographic language models.

The most prominent categories across most tokenizers were "spaced alpha" and "inner alpha" tokens, representing full words or word beginnings, and word parts or suffixes, respectively. However, this distribution showed notable differences for tokenizers designed for ideographic languages.

\textbf{Spaced alpha tokens:} These constituted the largest proportion for most tokenizers, ranging from approximately 30\% to over 60\%. The falcon tokenizer showed the highest percentage at over 60\%.

\textbf{Inner alpha tokens:} Generally ranging from about 30\% to 50\%, these formed the second largest category for most tokenizers.

\textbf{Ideographic language tokenizers:} The yi-34b (AI et al., 2024) and Qwen tokenizers, both designed for Chinese language models, showed distinct patterns:  Yi-34b had an unusually high proportion of inner alpha tokens (over 50\%) and a lower proportion of spaced alpha tokens;  qwen demonstrated lower percentages for both spaced alpha and inner alpha compared to other tokenizers. These differences likely reflect the unique characteristics of Chinese and other ideographic writing systems, which do not use spaces between words and rely heavily on character combinations to convey meaning.

\textbf{Control tokens}: While generally low in percentage, these showed some variation across tokenizers, with mixtral having the highest proportion.

\textbf{Pure unicode, char errors, spaced errors, and inner errors}: These categories consistently represented less than 5\% of tokens across all tokenizers, indicating relatively low rates of encoding issues or errors.

\textbf{Inner other}: The dbrx (The Mosaic Research Team, 2024) tokenizer notably had a higher percentage in this category compared to other tokenizers. Interestingly, the qwen tokenizer also showed a relatively high proportion in this category, which may be related to its handling of Chinese characters and symbols.




The observed variations in token distribution across tokenizers have significant implications for their performance in multilingual contexts, particularly for ideographic languages. In terms of language representation, tokenizers with higher proportions of spaced alpha tokens, such as falcon, may be more efficient in representing languages that rely heavily on whole words. Conversely, tokenizers designed for ideographic languages, such as yi-34b and qwen, show distinct patterns that likely reflect the structure of languages like Chinese. The high proportion of inner alpha tokens in yi-34b and the unique distribution in qwen may be optimized for representing character combinations and contextual meanings crucial in ideographic writing systems. Regarding error handling, the consistently low percentages of error categories across tokenizers suggest generally robust encoding practices, even for complex writing systems like Chinese. The unique distribution seen in tokenizers like dbrx and qwen, particularly in the "inner other" category, may reflect their ability to handle non-alphabetic scripts or specialized notation, which is particularly important for ideographic languages. The more evenly distributed categories in the Qtok unified tokenizer suggest it may offer a versatile approach for multilingual applications, potentially at the cost of specialization for specific language families or writing systems. These results highlight the importance of tokenizer choice in multilingual LLM development, especially when considering ideographic languages. The variations observed could explain differences in model performance across languages and text types, emphasizing the need for careful tokenizer selection or custom development based on the intended application and language scope of an LLM.

The distinct patterns observed in yi-34b and qwen tokenizers underscore the challenges and specialized approaches required for effectively representing ideographic languages in LLMs. These findings suggest that future development of multilingual models may benefit from incorporating strategies that can effectively bridge the gap between alphabetic and ideographic writing systems.

\subsection{Alphabet-based Token Annotation and Classification}

Different encoding schemes are employed to represent the various characters used across natural languages. A single encoding system may encompass multiple natural languages, which are often part of the same language family. For instance, in the ISO/IEC 8859 series, the following encodings are utilized for different natural languages (ISO, 2020):

\begin{itemize}
    \item \textbf{ISO 8859-1 (Latin-1)}: Western European languages.
    \item \textbf{ISO 8859-2 (Latin-2)}: Central and Eastern European languages.
    \item \textbf{ISO 8859-3 (Latin-3)}: South European languages.
    \item \textbf{ISO 8859-4 (Latin-4)}: North European languages.
    \item \textbf{ISO 8859-5}: Cyrillic alphabet.
    \item \textbf{ISO 8859-6}: Arabic alphabet.
    \item \textbf{ISO 8859-7}: Greek alphabet.
    \item \textbf{ISO 8859-8}: Hebrew alphabet.
    \item \textbf{ISO 8859-9 (Latin-5)}: Turkish.
    \item \textbf{ISO 8859-10 (Latin-6)}: Nordic languages.
    \item \textbf{ISO 8859-11}: Thai.
    \item \textbf{ISO 8859-13 (Latin-7)}: Baltic languages.
    \item \textbf{ISO 8859-14 (Latin-8)}: Celtic languages.
    \item \textbf{ISO 8859-16 (Latin-10)}: Southeastern European languages.
\end{itemize}

\begin{figure}[H]
    \centering
    \includegraphics[width=1\linewidth]{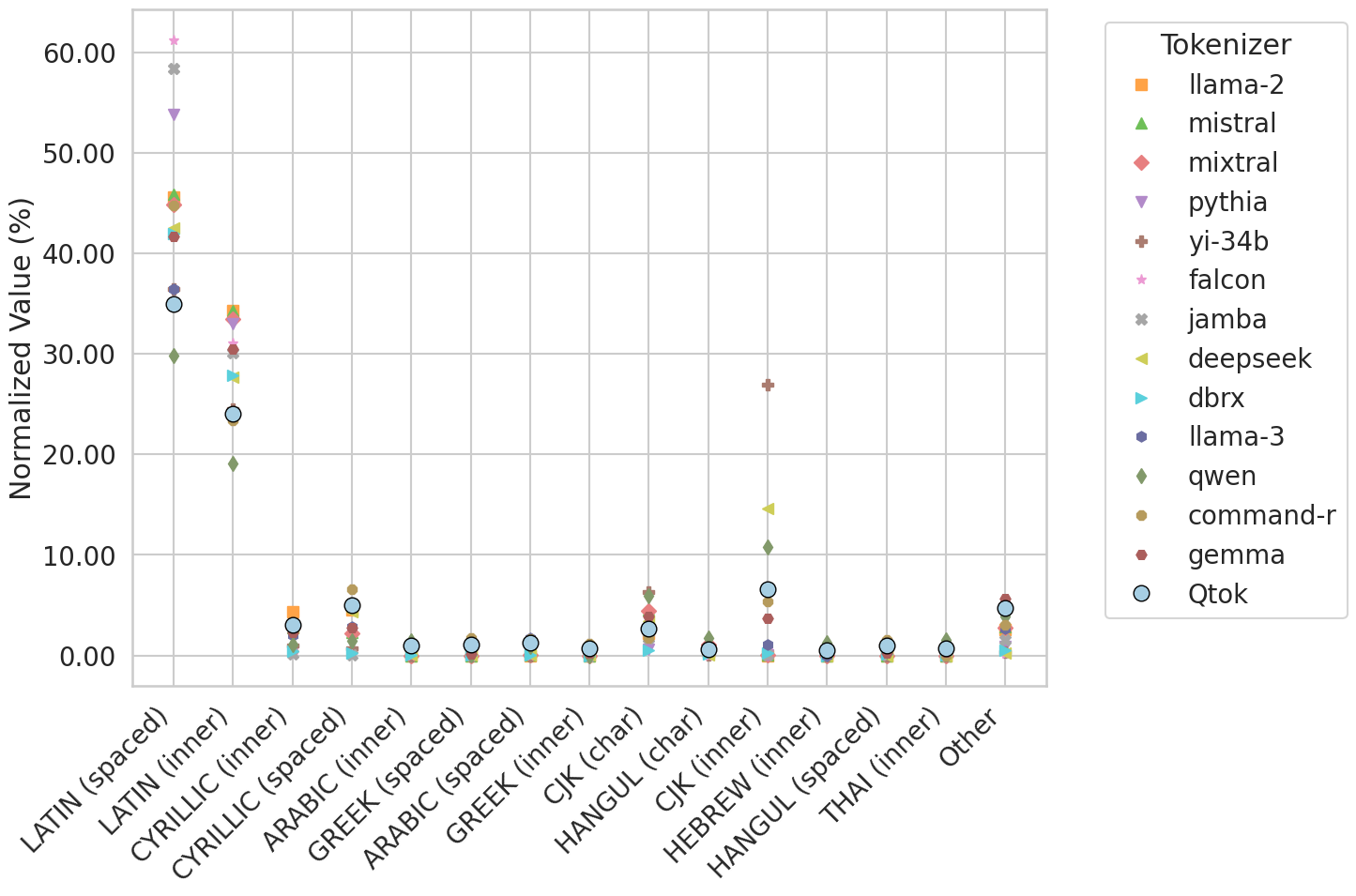}
    \caption{Unicode Range Representation in 13 Tokenizers. The joined tokenizer is provided as a reference point (Qtok).}
    \label{fig:enter-label6}
\end{figure}

In more than half of the encodings, multiple languages from the same language group (or family) are represented. The alphabets of such natural languages are similar, which highlights the challenge of identifying the specific language within a single encoding. This issue is particularly prominent with shorter tokens. For these tokens, determining the language can only be done reliably if unique linguistic characters are present within the token itself, allowing for a clear assignment to a particular language.

In the absence of such unique characters, short tokens may be attributed to several languages within the same group, as they might represent suffixes, articles, or word fragments that exist across multiple languages in the group. For longer tokens, which form whole words (or word lemmas), language identification becomes somewhat easier since the number of tokens that truly belong to multiple languages within the group is smaller than for shorter tokens. Moreover, for such tokens, language identification by humans is possible, meaning that machine learning methods can be effectively applied for detecting and identifying the language. 

However, current systems display a notable bias in language detection for both short and long tokens, impacting the accuracy of identifying the correct language across different token lengths.

The token classification process is divided into two stages: language classification and natural language classification within each language group.

\subsection{Language Classification}

The language classification process applies to all groups of natural language tokens discussed in the paper (internal tokens, space-start tokens, and character tokens). The classification process involves the following steps:

\paragraph{1.}We obtain the unified language form for each token by processing it character-by-character using the unicodedata module. In this context, the “unified form” refers to the specific Unicode encoding of each token, where we focus on the first word in its classification. For example, the Unicode encoding for the letter “A” is LATIN CAPITAL LETTER A. In our approach, we consider only the term LATIN and check for the presence of the word LETTER to avoid false positives from non-letter symbols. This method ensures that only actual letters are retained for further analysis, reducing noise from other types of characters.
\paragraph{2.}After obtaining the unified form, we classify it by the encodings used for representing natural languages. The languages identified in step 1 are then compared against a manually curated list of languages found in tokenizers designed for natural language representation. This comparison allows us to assess how well each tokenizer aligns with the languages it is intended to support, ensuring that the tokens are correctly mapped to the respective natural languages. 

We used the following languages: ‘ADLAM’, ‘CHAKMA’, ‘OL’, ‘BAMUM’, ‘OGHAM’, ‘BOPOMOFO’, ‘COPTIC’, ‘RUNIC’, ‘HALFWIDTH’, ‘MODI’, ‘KAITHI’, ‘LEPCHA’, ‘SHAVIAN’, ‘LIMBU’, ‘BATAK’, ‘PHOENICIAN’, ‘GLAGOLITIC’, ‘MANDAIC’, ‘BALINESE’, ‘SAMARITAN’, ‘PHAGS-PA’, ‘OLD’, ‘BLACK-LETTER’, ‘SUNDANESE’, ‘INSCRIPTIONAL’, ‘LISU’, ‘CHAM’, ‘TAGALOG’, ‘DESERET’, ‘TAGBANWA’, ‘BUGINESE’, ‘THAANA’, ‘MONGOLIAN’, ‘DINGBAT’, ‘JAVANESE’, ‘EGYPTIAN’, ‘GEORGIAN’, ‘NKO’, ‘TIFINAGH’, ‘GURMUKHI’, ‘BENGALI’, ‘SINHALA’, ‘ORIYA’, ‘TAI’, ‘KANGXI’, ‘CANADIAN’, ‘CHEROKEE’, ‘LAO’, ‘TELUGU’, ‘SYRIAC’, ‘TAMIL’, ‘BRAILLE’, ‘ETHIOPIC’, ‘MYANMAR’, ‘HEBREW’, ‘ARABIC’, ‘TIBETAN’, ‘HIRAGANA’, ‘CYRILLIC’, ‘GREEK’, ‘CJK’, ‘LATIN’, ‘KATAKANA’, ‘KHMER’, ‘THAI’, ‘ARMENIAN’, ‘YI’, ‘HANGUL’, ‘GOTHIC’, ‘MALAYALAM’, ‘DEVANAGARI’, ‘GUJARATI’, ‘KANNADA’, ‘MEETEI’, and ‘ARABIC-INDIC’.

\begin{table}[h]
    \centering
    \caption{Unicode ranges with >1\% representation in 13 tokenizers and Qtok joined the tokenizer.}
    \begin{tabularx}{\textwidth}{lXXXXXXXXXXXXXXXXXX}
        \toprule
        \textbf{Category} & \rotatebox{90}{\textbf{Qtok}} & \rotatebox{90}{\textbf{llama-2}} & \rotatebox{90}{\textbf{mistral}} & \rotatebox{90}{\textbf{mixtral}} & \rotatebox{90}{\textbf{pythia}} & \rotatebox{90}{\textbf{yi-34b}} & \rotatebox{90}{\textbf{falcon}} & \rotatebox{90}{\textbf{jamba}} & \rotatebox{90}{\textbf{deepseek}} & \rotatebox{90}{\textbf{dbrx}} & \rotatebox{90}{\textbf{llama-3}} & \rotatebox{90}{\textbf{qwen}} & \rotatebox{90}{\textbf{command-r}} & \rotatebox{90}{\textbf{gemma}} \\
        \midrule
        LATIN (inner)       & 24.01 & 34.28 & 34.28 & 33.48 & 33.03 & 24.47 & 31.06 & 30.15 & 27.68 & 27.83 & 23.48 & 19.08 & 23.34 & 30.48 \\
        LATIN (spaced)      & 34.98 & 45.62 & 45.91 & 44.84 & 53.83 & 36.50 & 61.23 & 58.35 & 42.56 & 41.94 & 36.50 & 29.86 & 44.80 & 41.66 \\
        CYRILLIC (inner)    & 3.05  & 4.38  & 2.80  & 2.73  & 0.44  & 0.97  & 0.03  & 0.12  & 2.75  & 0.40  & 2.05  & 1.12  & 3.25  & 2.26  \\
        CYRILLIC (spaced)   & 5.03  & 4.53  & 2.28  & 2.23  & 0.22  & 0.73  & 0.02  & 0.06  & 4.35  & 0.27  & 2.89  & 1.46  & 6.54  & 2.78  \\
        HANGUL (char)       & 0.61  & 0.35  & 1.08  & 1.06  & 0.04  & 0.04  & 0     & 0.46  & 0.01  & 0.13  & 0.47  & 1.72  & 0.31  & 0.66  \\
        CJK (char)          & 2.67  & 2.19  & 4.55  & 4.44  & 0.60  & 6.28  & 1.65  & 1.69  & 3.42  & 0.55  & 1.80  & 5.94  & 1.73  & 3.84  \\
        GREEK (inner)       & 0.73  & 0     & 0     & 0     & 0.16  & 0     & 0     & 0     & 0.03  & 0     & 0.65  & 0     & 1.19  & 0.27  \\
        ARABIC (inner)      & 1.02  & 0     & 0     & 0     & 0.04  & 0     & 0     & 0.67  & 0.01  & 0.03  & 1.08  & 1.41  & 0.98  & 0.90  \\
        ARABIC (spaced)     & 1.26  & 0     & 0.01  & 0.01  & 0.06  & 0     & 0     & 0.54  & 0.01  & 0.03  & 1.73  & 0.88  & 1.59  & 1.38  \\
        GREEK (spaced)      & 1.03  & 0     & 0     & 0     & 0.13  & 0     & 0.01  & 0.01  & 0.02  & 0.01  & 0.37  & 0.01  & 1.75  & 0.17  \\
        CJK (inner)         & 6.57  & 0     & 0.01  & 0.01  & 0.02  & 26.98 & 0.57  & 0.08  & 14.58 & 0.20  & 1.03  & 10.80 & 5.35  & 3.70  \\
        HEBREW (inner)      & 0.53  & 0     & 0     & 0     & 0     & 0     & 0     & 0.57  & 0     & 0     & 0     & 1.22  & 0.22  & 0.26  \\
        THAI (inner)        & 0.72  & 0     & 0     & 0     & 0     & 0     & 0     & 0     & 0     & 0.02  & 0.88  & 1.58  & 0     & 0.37  \\
        HANGUL (spaced)     & 0.94  & 0     & 0     & 0     & 0     & 0     & 0     & 0.01  & 0     & 0.13  & 0.88  & 0.44  & 1.54  & 0.21  \\
        Other               & 4.71  & 2.54  & 2.83  & 2.77  & 0.86  & 0.35  & 0.17  & 1.51  & 0.21  & 0.52  & 2.69  & 3.89  & 3.08  & 5.62  \\
        \bottomrule
    \end{tabularx}
\end{table}

\subsection{Natural language classification}

Classification of natural language, unlike script classification, is a much more complex process. Currently, there is a challenge in associating a token with a specific language: words and even entire phrases can appear in multiple languages, not to mention shorter tokens (those shorter than the average word length) and lemmas. Consequently, classifying a token to a single language remains a complex and unresolved issue in the modern world. In our work, we propose a partial solution to this problem using the \textit{langid.py} tool (Lui \& Baldwin, 2012):

The choice of this tool is driven by its classification quality and the ability to rank natural languages for each token. The improved process follows the algorithm outlined below:

\paragraph{1.}Obtaining the scripts for each token using the classification method described in section 2.5.
\paragraph{2.}Identification of permissible natural language codes. Each script is mapped to a list of natural language codes following the ISO 639-1 format. The choice of this format is driven by the selection of \textit{langid.py} as the tool for natural language classification, as \textit{langid.py} specifically utilizes the ISO 639-1 format.
\paragraph{3.}Classification of natural languages for the token from the permissible natural language codes is performed by ranking and selecting the most likely candidates. The selection is based on the following rule: the number of top languages considered varies depending on the token length. \\ 
\begin{itemize}
    \item 5 languages for 1-character tokens
    \item 4 languages for 2-character tokens
    \item 3 languages for 3-4 character tokens
    \item 2 languages for 5 characters tokens
    \item 1 language for tokens 6 characters or longer
\end{itemize}

We chose the number of languages linked to a token empirically – according to our observations, the bias of the standard classification in the \textit{langid.py} library for tokens of length 6 and above is significantly lower than the bias for determining the language of 5-character tokens. In turn, the bias for 5-character tokens is lower than for tokens of 3-4 characters. Tokens of length 2 have the highest number of potential languages, as they are the hardest to assign to any single specific language.
\begin{figure}[H]
    \centering
    \includegraphics[width=1\linewidth]{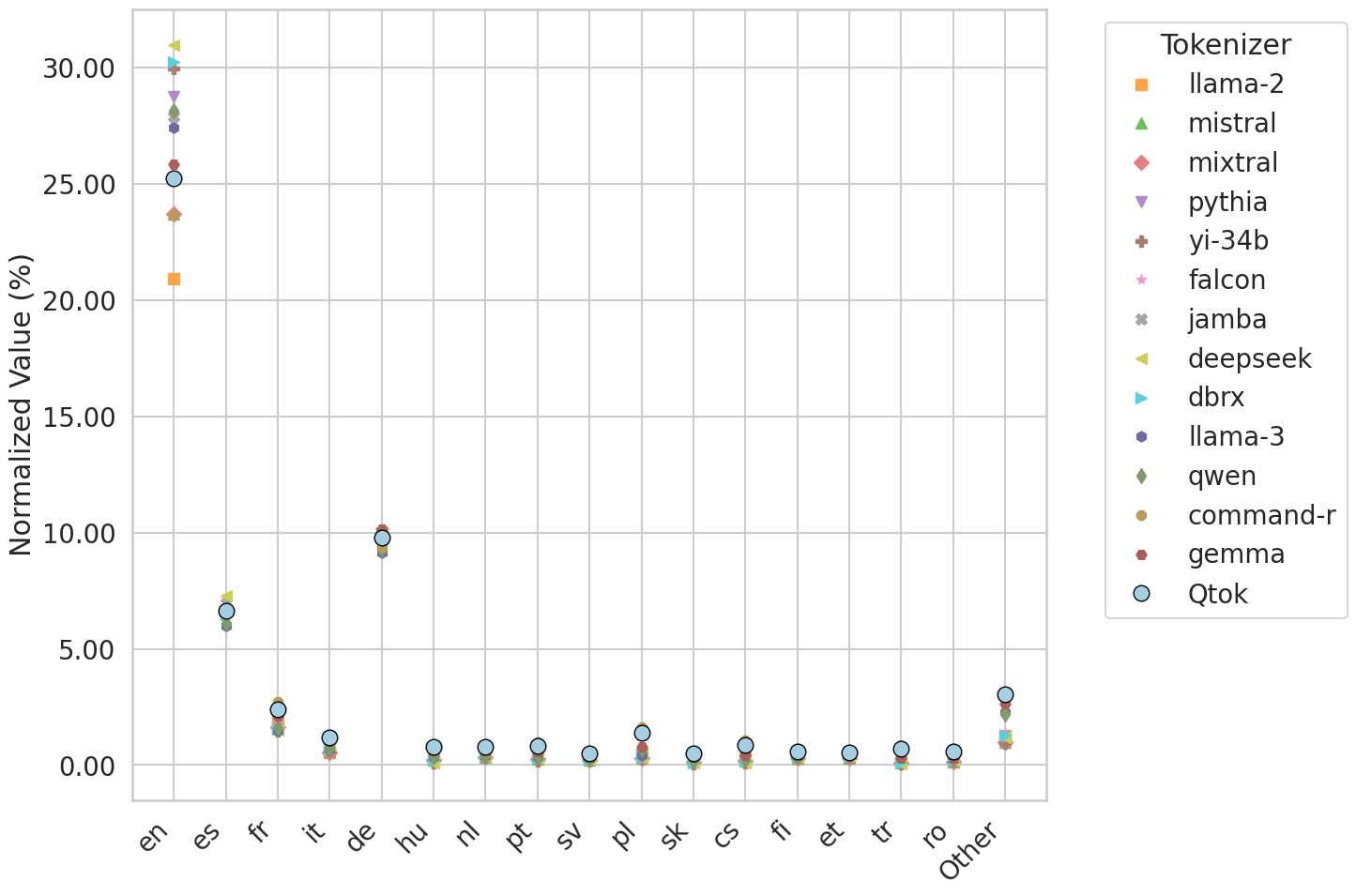}
    \caption{Language detection in latin unicode range in 13 Tokenizers. The joined tokenizer is provided as a reference point (Qtok).}
    \label{fig:enter-label7}
\end{figure}

\begin{figure}[H]
    \centering
    \includegraphics[width=1\linewidth]{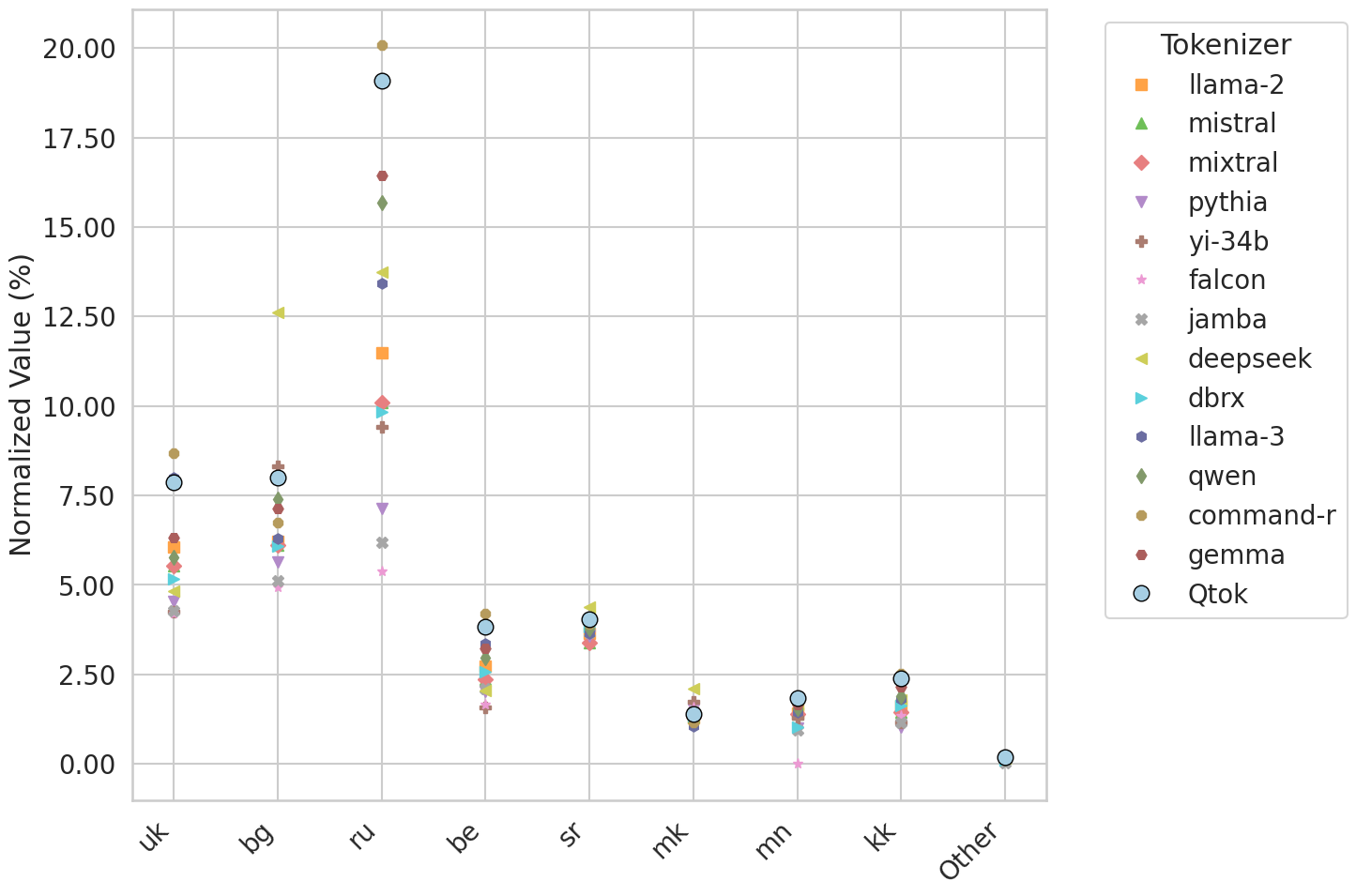}
    \caption{Language detection in cyrillic unicode rangein 13 Tokenizers. The joined tokenizer is provided as a reference point (Qtok).}
    \label{fig:enter-label8}
\end{figure}
However, this approach is still not robust and does not address most issues inherent in machine classification, although it partially reduces bias when classifying short tokens. 

\subsection{Core tokens between tokenizers}

In this study, tokenizers are considered identical if they share the exact same vocabulary. An analysis of core tokens across various tokenizer groups reveals distinct patterns based on vocabulary size, highlighting the diversity and specialization of tokenization strategies in contemporary language models.

\subsubsection{Grouping Tokenizers by Vocabulary Size}

We grouped tokenizers based on their vocabulary size. However, the vocabulary sizes of most tokenizers vary significantly. If we were to group tokenizers with identical vocabulary sizes, as mentioned earlier, the groups would consist of only identical tokenizers, introducing substantial bias into the metric values for these groups.

During the analysis, we observed that the vocabulary sizes of most tokenizers are multiples of powers of two. Therefore, we decided to group the tokenizers by powers of two. Tokenizers with vocabulary sizes not exactly equal to a power of two were assigned to the nearest corresponding group (e.g., a tokenizer with a vocabulary of 50,000 tokens was placed in the group with 65,000 tokens).
As a result, we identified four groups of models:

\begin{enumerate}
    \item \textbf{Models with Approximately 32,000 Vocabulary Tokens:} \\
    This group comprises 21 models, each featuring a vocabulary size of roughly 32,000 tokens. Collectively, these models encompass approximately 40,358 unique tokens. Of these, 23,957 are identified as core tokens, which are consistently present across multiple tokenizers within this group. Additionally, there are 54 singleton tokens—tokens that appear exclusively in a single tokenizer. These models are positioned in the top-left quadrant of the token group distribution, indicating a concentration of models with smaller, more standardized vocabularies.
    \item \textbf{Models with Vocabulary Sizes Between 50,000 and 64,000:} \\
    This group consists of 12 models with vocabulary sizes ranging from 50,000 to 64,000 tokens. The group includes approximately 108,934 unique tokens, with 31,707 core tokens shared among the tokenizers. The presence of 8,514 singleton tokens in this group deviates from the mode observed across all groups. This deviation is primarily due to the model families within this group, which includes four families: the Pythia family with 7 models, the Yi-34b family with 2 models, the Jamba family with 2 models, and the Falcon family with 1 model. Notably, the Falcon family, consisting of only one model, significantly contributes to this deviation from the mode.
    \item \textbf{Models with Vocabulary Sizes Between 100,000 and 150,000:} \\
    Sixteen models in this analysis possess vocabularies ranging from 100,000 to 150,000 tokens. These models collectively include 198,487 unique tokens, of which 53,243 are core tokens shared across the group. This category contains 22 singleton tokens, reflecting some unique vocabulary elements but generally high commonality within the group.
    \item \textbf{Models with Approximately 256,000 Vocabulary Tokens:} \\
    Nine models in this group have vocabularies around 256,000 tokens each. These models collectively feature 359,390 unique tokens, including 149,579 core tokens and no singleton tokens, indicating a high level of vocabulary overlap and standardization among these large-vocabulary models.
\end{enumerate}

\begin{table}[h]
    \centering
    \caption{Tokenizers grouped by vocabulary size magnitude}
    \begin{tabularx}{\textwidth}{l X X X X X}
        \toprule
        \textbf{\# Models} & \textbf{\# Distinct Tokenizers} & \textbf{Vocab Size} & \textbf{Total Tokens in Group} & \textbf{Core Tokens} & \textbf{Unique Tokens} \\
        \midrule
        21  & 3 & 32,000 & 40,358  & 23,957  & 54   \\
        12  & 4 & 32,000-66,000 & 108,934 & 31,707  & 8,514 \\
        16  & 4 & 100,000-152,000 & 198,487 & 53,243  & 22   \\
        9   & 2 & \textasciitilde256,000 & 359,390 & 149,579 & 0    \\
        \bottomrule
    \end{tabularx}
\end{table}

\subsubsection{Small number of unique tokens}
In three of the four tokenizer groups, the number of singleton tokens is relatively small compared to the vocabulary size of the models. This is because these tokenizer groups do not include model families consisting of only a single model. \\ \\
The non-zero count of singleton tokens in two of the four model groups appears unusual, as models within the same family typically share identical tokenizers. However, the observed differences in seemingly identical tokenizers stem from reserved tokens. For instance, two models sharing the same tokenizer may have different reserved (system) tokens, which results in conditional differences between these otherwise identical tokenizers. \\ \\
For example, in the 100,000–152,000 vocab size token group, the singleton tokens are as follows: \\ \\
$'\langle \| \_unused\_0\_ \| \rangle '$, 
$'\langle \| endoftext \| \rangle '$, 
$'\langle \| fim\_prefix \| \rangle '$, 
$'\langle \| fim\_middle \| \rangle '$, 
$'\langle \| fim\_suffix \| \rangle '$, 
$'\langle \| \_unused\_1\_ \| \rangle '$, 
$'\langle \| \_unused\_2\_ \| \rangle '$, 
$'\langle \| \_unused\_3\_ \| \rangle '$, 
$'\langle \| \_unused\_4\_ \| \rangle '$, 
$'\langle \| \_unused\_5\_ \| \rangle '$, 
$'\langle \| \_unused\_6\_ \| \rangle '$, 
$'\langle \| \_unused\_7\_ \| \rangle '$, 
$'\langle \| \_unused\_8\_ \| \rangle '$, 
$'\langle \| \_unused\_9\_ \| \rangle '$, 
$'\langle \| \_unused\_10\_ \| \rangle '$, 
$'\langle \| \_unused\_11\_ \| \rangle '$, 
$'\langle \| \_unused\_12\_ \| \rangle '$, 
$'\langle \| \_unused\_13\_ \| \rangle '$, 
$'\langle \| \_unused\_14\_ \| \rangle '$, 
$'\langle \| \_unused\_15\_ \| \rangle '$, 
$'\langle \| endofprompt \| \rangle '$, 
$'\langle \| pad \| \rangle '$

Meanwhile, the 32,000 vocab size token group contains the following singleton tokens:

'[IMG]', '[PREFIX]', '[MIDDLE]', '[SUFFIX]', '[IMG\_BREAK]', '[IMG\_END]', '[REFERENCE\_DOC\_19]', '[REFERENCE\_DOC\_18]', '[REFERENCE\_DOC\_17]', '[REFERENCE\_DOC\_16]', '[REFERENCE\_DOC\_15]', '[REFERENCE\_DOC\_14]', '[REFERENCE\_DOC\_13]', '[REFERENCE\_DOC\_12]', '[REFERENCE\_DOC\_11]', '[REFERENCE\_DOC\_10]', '[REFERENCE\_DOC\_9]', '[REFERENCE\_DOC\_8]', '[REFERENCE\_DOC\_7]', '[REFERENCE\_DOC\_6]', '[REFERENCE\_DOC\_5]', '[REFERENCE\_DOC\_4]', '[REFERENCE\_DOC\_3]', '[REFERENCE\_DOC\_2]', '[REFERENCE\_DOC\_1]', '[REFERENCE\_DOC\_0]', '[control\_8]', '[control\_9]', '[control\_10]', '[control\_11]', '[control\_12]', '[control\_13]', '[control\_749]', '[control\_750]', '[control\_751]', '[control\_752]', '[control\_753]', '[control\_754]', '[control\_755]', '[control\_756]', '[control\_757]', '[control\_758]', '[control\_759]', '[control\_760]', '[control\_761]', '[control\_762]', '[control\_763]', '[control\_764]', '[control\_765]', '[control\_766]', '[control\_767]', '[control\_768]', '$\langle \| im\_start \| \rangle$'

As shown, all tokens except one, consisting of 16 spaces, are reserved tokens specific to the models.

\subsection{Qtok Evaluation Metrics}

In this study, we suggested a comprehensive set of metrics to evaluate tokenizer quality. These metrics were designed to assess various aspects of tokenizer performance, including uniqueness, core token coverage, and language representation. 

Our analysis began with fundamental tokenizer attributes including tokenizer size from configuration file (it can be different from actual size), tokenizer family, and real tokenizer size. These metrics provide insight into the basic structure and capacity of each tokenizer. We then examined the tokenizer's coverage by quantifying the number of tokens found in Qtok (joined tokenizer) and the number of unseen tokens, which helps evaluate the tokenizer's ability to handle diverse text inputs.

To assess language representation, we evaluate  the tokenizer's vocabulary across various character and token types. This included metrics for alphabetic characters (both standalone and within words), numbers, and coding tokens from programming languages. We also evaluated the tokenizer's handling of non-alphabetic characters and spaces, providing a comprehensive view of its ability to process different text elements (inner, space start, single char).

The evaluation extended to more subtlety aspects of tokenization, such as the handling of uneven bytes and unparsable characters or sequences. These metrics offer insights into the tokenizer's robustness when faced with unusual or potentially problematic inputs. By including control tokens in our analysis, we also considered the tokenizer's capability to manage special tokens that may be crucial for certain instruct models.

Through this multistep evaluation approach, we aim to provide a thorough understanding of each tokenizer's oddities. This comprehensive set of metrics allows for detailed comparisons between different tokenizers, enabling researchers to make more informed decisions when selecting a tokenizer for specific applications or when developing new tokenization strategies.

\subsection{Qtok the tool for tokenizers quality control}

Our analysis of various tokenizer metrics reveals significant differences across different tokenization approaches, potentially reflecting diverse design philosophies and intended use cases. The metrics examined include control tokens, pure unicode, character types (alpha and other), spacing, and error handling.

In terms of control tokens, we observe a wide range, with jamba having the highest count at 1549 and pythia the lowest at just 4. This substantial difference might be attributed to the specific requirements of instruction fine-tuning.

\begin{table}[ht]
\caption{Tokenizer Metrics: Light blue indicates minimum values or deviations from 256 in "Pure Unicode." Light yellow shows maximum values, and light green highlights Qtok. \\}
\centering
\renewcommand{\arraystretch}{1.2}
\begin{adjustbox}{width=1\textwidth}
\begin{tabular}{|l|r|r|r|r|r|r|r|r|r|r|r|r|}
\hline
\textbf{Tokenizer} & \textbf{Control Tokens} & \textbf{Pure Unicode} & \textbf{Char Alpha} & \textbf{Spaced Alpha} & \textbf{Inner Alpha} & \textbf{Char Other} & \textbf{Spaced Other} & \textbf{Inner Other} & \textbf{Unicode Flanks} & \textbf{Char Errors} & \textbf{Spaced Errors} & \textbf{Inner Errors} \\
\hline
\rowcolor{lightgreen} 
Qtok & 2991 & 256 & 19166 & 196904 & 168797 & 4091 & 3617 & 30902 & 5885 & 916 & 827 & 3554 \\
llama-2 & 16 & 256 & 1629 & 16054 & 12375 & 257 & 349 & 1039 & \cellcolor{lightblue}0 & \cellcolor{lightblue}13 & 5 & 18 \\
mistral & 8 & 256 & 2710 & \cellcolor{lightblue}15425 & \cellcolor{lightblue}11870 & 290 & 325 & \cellcolor{lightblue}1001 & \cellcolor{lightblue}0 & 67 & 11 & 39 \\
mixtral & 773 & 256 & 2710 & \cellcolor{lightblue}15425 & \cellcolor{lightblue}11870 & 290 & 325 & \cellcolor{lightblue}1002 & \cellcolor{lightblue}0 & 67 & 11 & 39 \\
pythia & \cellcolor{lightblue}4 & \cellcolor{lightblue}185 & \cellcolor{lightblue}690 & 27288 & 17000 & 63 & 1463 & 2805 & 486 & 65 & 33 & 175 \\
yi-34b & 195 & 256 & 4204 & 23899 & 33550 & 107 & 344 & 1412 & \cellcolor{lightblue}0 & \cellcolor{lightblue}13 & \cellcolor{lightblue}3 & \cellcolor{lightblue}9 \\
falcon & 7 & \cellcolor{lightblue}170 & 1194 & 39831 & 20581 & \cellcolor{lightblue}58 & \cellcolor{lightblue}53 & 2039 & 946 & 19 & 20 & 106 \\
jamba & \cellcolor{lightyellow}1549 & 256 & 2051 & 38996 & 20709 & 297 & 401 & 1176 & \cellcolor{lightblue}0 & 42 & 13 & 46 \\
deepseek & 26 & \cellcolor{lightblue}184 & 3660 & 46954 & 45059 & 113 & 711 & 2390 & 558 & 36 & 8 & 319 \\
dbrx & 33 & \cellcolor{lightblue}184 & 1040 & 42603 & 28628 & 96 & 1190 & 22202 & 782 & 35 & 640 & 2847 \\
llama-3 & 265 & \cellcolor{lightblue}211 & 3597 & 55444 & 39084 & 210 & 1314 & \cellcolor{lightyellow}23131 & 1361 & 66 & \cellcolor{lightyellow}659 & \cellcolor{lightyellow}2914 \\
qwen & 15 & 256 & 14597 & 50981 & 54914 & 2915 & 1198 & 21791 & 1457 & 35 & 640 & 2847 \\
command-r & 48 & \cellcolor{lightblue}198 & 5899 & \cellcolor{lightyellow}145978 & 92155 & 182 & 1121 & 4546 & \cellcolor{lightyellow}4489 & 33 & 92 & 288 \\
gemma & 151 & \cellcolor{lightblue}255 & \cellcolor{lightyellow}15712 & 123481 & \cellcolor{lightyellow}102881 & \cellcolor{lightyellow}3841 & \cellcolor{lightyellow}1698 & 6728 & \cellcolor{lightblue}0 & \cellcolor{lightyellow}913 & 81 & 289 \\
\hline
\end{tabular}
\end{adjustbox}

\label{tab:tokenizer_metrics}
\end{table}

For pure unicode handling, we see a clear division: llama-2, mistral, mixtral, yi-34b, jamba, and qwen all show an absolute count of 256, while the rest demonstrate varying non-absolute counts. The absence of 256 Unicode characters in some tokenizers suggests that not all Unicode symbols appear in the dataset in sufficient quantities, meaning they may not be encountered during tokenization. When we observe exactly 256 characters, it’s likely that this range was manually inserted to ensure its presence.

Character handling metrics show interesting patterns. Gemma consistently demonstrates high counts across various character-related metrics, including char alpha (15712), inner alpha (102881), char other (3841), and spaced other (1698). In contrast, pythia shows the lowest count for char alpha (690), while falcon has the lowest counts for char other (58) and spaced other (53). The low counts for falcon might be attributed to its focus on the Arabic language, potentially resulting in a training dataset with fewer non-alphabetic characters compared to other models.

Spacing-related metrics reveal that command-r has the highest count for spaced alpha (145978), while mistral and mixtral share the lowest (15425). The number of tokens that begin with a space is often indicative of whether the tokenizer is truly BPE (Byte Pair Encoding) or if it follows a different tokenization method. This also depends on how much the language has been influenced by English. The more English present in the dataset, the more tokens we expect to begin with a space. In languages that are less represented, we are likely to see a growing number of tokens that start with a space and represent internal parts of words, such as prefixes, roots, and suffixes.

Additionally, for Chinese, there is typically a low number of tokens that start with spaces, as the language itself generally lacks spaces. This issue with spaces is also observed in languages where spaces do not serve as delimiters between words.

Gemma has the highest char errors count (913), while llama-3 leads in spaced errors (659) and inner errors (2914). On the other end of the spectrum, yi-34b consistently shows low error counts across all three error-related metrics The highest number of single-character encoding errors is observed in Gemma. It appears that part of Gemma’s dataset was enriched by texts with broken encoding, and the dataset was large enough that individual UTF symbols and other encodings formed separate tokens.

Other errors consist mainly of unusual tokens with special characters, such as combinations of newlines, tabs, spaces, and brackets. These are likely symbols from programming languages with broken encoding, which resulted in the formation of a large number of anomalous tokens. The full list of these tokens are available in the github repository.

While our analysis provides insights into the varying approaches of different tokenizers, many of the observed differences require further investigation to fully understand their underlying causes. Factors such as training data composition, specific language focus, and intended application of each tokenizer likely play crucial roles in shaping these metrics.

\subsection{Qtok evaluation for quality control for  tokenizers}

Finally, we decided to examine closed models and compare how they differ from others. Initially, we had not planned to include them, but this omission would be significant. These models should not be overlooked, as they are among the most popular as of September 2024.

However, due to the closed nature of these models, official tokenizers are inaccessible. As a result, we relied on unofficial tokenizers, sourced from community contributions, which were obtained through various non-standard methods. While these unofficial tokenizers provide a practical alternative, they come with potential variability in quality. Nevertheless, our analysis of tokenizers for models such as GPT-4o (Lochner, 2024), Grok (Lochner, 2023), and Claude (Claude-v1-Tokenizer, 2024), using primary qtok metrics, revealed no statistically significant deviations



\section{Discussion}

\subsection{Tokenizer availability problem}

There is a significant issue with the accessibility of tokenizers, as proprietary and closed models often do not publicly release their tokenizers. Some repositories claim to contain tokenizers, but as far as we understand, these tokenizers are often extracted from various tools providing closed solutions, making them unreliable. While we did not exclude such tokenizers from our analysis, the problem of accessing tokenizers for a comprehensive comparison remains unresolved.

There are many models where accessing the tokenizer is quite challenging, and this problem is multi-layered. It’s difficult to obtain the tokenizer itself, and even if one is available, it’s unclear which dataset was used to train it. Datasets specifically for training tokenizers are generally not released. However, it is evident that the tokenizer was likely trained on a smaller dataset, different from the one used for model training. We now suspect that the dataset used for training the model and the dataset for the tokenizer may differ.

Many common model errors can actually be traced back to tokenizer issues, or rather, tokenizer characteristics. For example, a frequent issue observed in models as of September 2024 involves difficulties in counting the number of letters in a word or correctly comparing two numbers separated by a decimal point. The model often misinterprets them as two distinct numbers separated by a period, rather than as a single number with decimal places.

Moreover, the tokenizer is a distinctive feature of a model, and by analyzing how it tokenizes certain words, it is often possible to identify the underlying model, even if the authors do not disclose the exact details. Due to the intense competition in model development, many attempt to fine-tune existing foundational models and present them as entirely new ones. However, the true nature of the model can often be inferred based on the tokenizer it uses internally.

\subsection{Attempts to create unified core tokenizer}

Initially, we had the idea that by taking all the tokenizers, we would find a large number of common tokens among them. However, this approach didn’t work as expected, and we discovered that tokenizers across models differ significantly in size. As a result, creating a unified tokenizer from all of them wasn’t feasible. However, we simplified this concept by developing a unified tokenizer for models with the same size, which we successfully implemented. Nevertheless, our broader goal of creating a large, unified tokenizer across all models was not achieved as we had hoped.
Nonetheless, we managed to collect over 430,000 distinct tokens, which we have named the “unified tokenizer,” or Qtok. This could potentially serve as a benchmark for evaluating other tokenizers, particularly by assessing how many tokens in a new tokenizer are captured relative to Qtok. We expect that most tokens are already covered by existing tokenizers, though this assumption requires further investigation. It will be especially important to see how, when focusing on individual languages, tokenization enriches the dataset, particularly with tokens that begin with spaces.

\subsection{Tokenizer as latent representation of training tokenizer dataset}

Another important result of our study is the use of core tokens as a metric for determining how artificially enriched a dataset might be with specific language patterns. This offers a new way of evaluating the composition of a dataset and the extent to which it has been engineered to favor particular languages. Since BPE tokenizers build their vocabularies based on token frequency within a dataset, this metric allows us to measure the level of bias introduced during dataset creation. To our knowledge, this approach to analyzing tokenizers and datasets has not been widely explored.

It is important to emphasize once again that datasets used for training tokenizers are almost never publicly available (one of exceptions Kumar et al., 2024). We were unable to find any datasets specifically used for tokenizers. This discrepancy between the dataset used for the tokenizer and the one used for model training can have a significant impact on the model’s overall performance.

Moreover, we hypothesize that in the future, tokenizers could be used for reverse-engineering the dataset on which they were trained. This could work both ways: allowing us to identify weaknesses in the dataset and suggest improvements, as well as providing insights into how the tokenizer itself could be enhanced. Even if the tokenizer faced limitations due to the dataset, it might be possible to supplement it with additional tokens, as was done with r-command and smiles chars or with many models that incorporate the full range of 256 bytes.

Additionally, it has been demonstrated that tokens present in the tokenizer but nearly absent from the training dataset pose a potential security risk for the model. These so-called “dangling” tokens, as previously reported, can serve as vulnerabilities in the model’s operation (Land \& Bartolo, 2024; D. Wang et al., 2024).

One of the clearest insights from analyzing tokenizers is how the language of the dataset influences the tokenizer. A simple metric for this is the number of tokens that start with a space for a given language, which provides a good estimate of how much that language was represented in the original dataset used to train the tokenizer. On the flip side, this reveals issues with underrepresented languages, especially those outside of the most popular ones found in Common Crawl. It remains unclear where to source datasets for these languages, suggesting that we may need to address this issue by artificially enriching tokenizers with words from less-represented languages. In addition to language enrichment, the longer the tokens, the better represented the language is in the dataset. This indicates that the tokenizer captures more complex word structures, suggesting a more comprehensive representation of the language within the dataset.

\subsection{Manually curated tokenizers}

Given the challenges with datasets for tokenizers, an alternative approach to solving this issue could be the creation of manually-crafted tokenizers. This is particularly relevant for specialized tokenizers, such as those tailored to specific domains. A notable example is tokenizers for programming languages, where they need to be enriched with tokens characteristic of the particular programming language in which the model is expected to perform effectively. This approach could provide better results for domain-specific tasks, ensuring the tokenizer is well-optimized for that context.

Our Qtok is the first to provide a framework for analyzing tokenizers and evaluating their quality. As far as we know, no other such tools currently exist, and tokenizers remain one of the least studied components of the entire model training pipeline. While there has been significant focus on dataset quality, tokenizer quality has largely stayed out of the spotlight. However, tokenizers can substantially impact model performance. This is evident, for instance, in the case of the latest generation models, where one key to their success was a significant expansion of the tokenizer—even for models with relatively small sizes, with Gemma as an example of such improvements.

We were offered a set of metrics, but we recognize that we are just at the beginning of evaluating tokenizers’ quality. It’s likely that more sophisticated metrics will emerge beyond those we initially proposed. Nevertheless, the introduction of these metrics already provides significant opportunities for using our tool to compare how well a tokenizer performs, not just for English, but for other languages as well. Thank you for your attention!

Another crucial aspect of tokenizers is how well they tokenize words and, ultimately, their cost efficiency. One phenomenon we’ve observed is that models for English tend to be significantly cheaper than for less common languages, because rarer languages are tokenized into a larger number of tokens. Since model usage is typically billed per million tokens, the cost of performing the same task in English versus a rarer language can differ by several orders of magnitude. This creates a significant cost disparity for non-English languages, making the same operation much more expensive depending on the language.

Creating custom tokenizers is an economically beneficial task because it can significantly reduce the cost of running models on non-English languages for end users. We believe this presents a great opportunity for optimizing tokenizer performance. By simply adding the necessary tokens for a specific language to the tokenizer before training the model, substantial cost savings can be achieved. This approach allows for better efficiency in tokenization and ultimately lowers the operational costs for various languages.

Throughout our research, we observed the release of numerous new models, each bringing improvements or optimizations to specific components compared to existing models. Changes such as model size, the number of parameters, and overall model architecture—including optimizations for running models on less powerful devices—were consistently seen across various models. However, the introduction of new, improved tokenizers was notably absent. Developers seem to overlook the potential of modifying tokenizers. In our view, enhancing the tokenizer (through enrichment or other means) without altering the model architecture could lead to significant improvements in model performance.

It is important to note that a lot of the recent advancements in machine learning have been focused on models, specifically on model architecture and, more recently, on datasets. However, we continue to emphasize that the tokenizer is just as important—if not more so—than the dataset or model structure. Tokenizers play a fundamental role in how the model interprets and processes data, and improvements in tokenization can have a profound impact on overall model performance.

\subsection{Tokenizer diversity and language representation}

One of the most prominent issues with tokenizers is their handling of non-major languages. A surprising problem is the incomplete inclusion of alphabets from less commonly represented languages. For many tokenizers, it’s challenging to find full alphabets for languages spoken by tens of millions of people. Even when tokenizers are trained on datasets that include these languages, the representation is often so sparse that the alphabet of the language doesn’t make it into the tokenizer. This is particularly true for languages with a large number of unique characters, such as geographic languages, scripts with hieroglyphs, and Chinese. These gaps highlight the need for better tokenization strategies for languages with complex or less frequently represented writing systems.

One of our hypotheses was that if we start combining tokens with tokenizers, we would eventually reach a plateau (Figure 4). However, what we observed is that we are far from reaching such a plateau, and tokenizers can be expanded significantly. If we aim to have around 5,000 words in each tokenizer for every language, and we consider about 200-300 languages (totaling roughly 1.5 million tokens), we can estimate the lower boundary for a high-quality tokenizer that covers all major languages—at least those languages with significant digital content, such as those used in IT and available on Wikipedia. This calculation provides a baseline for creating a truly comprehensive tokenizer for multilingual applications.

\subsection{We Need Toolset for Minority Language Analysis in LLMs}

An unexpected problem we encountered is that even identifying which language a specific character belongs to is far from a trivial task, and it remains unsolved. There are no existing tools that can reliably determine the language of a character. At best, we can determine the Unicode range to which a character belongs, but that does not necessarily tell us which language it is used for. As a result, identifying the language of a character based purely on its symbol is not currently feasible. This limitation complicates efforts to develop better tokenizers, especially for multilingual applications.

There are many tools available for identifying the natural language of a given text, but these tools come with several significant issues:

\subsubsection{ML-Based System Bias}
\paragraph{Dramatic bias increases with shorter texts:} These tools are heavily biased when the length of the text decreases, and the problem is even more pronounced when dealing with individual tokens, which are much shorter than typical texts used for language classification.

\paragraph{Dataset problem:} It is often unclear on which texts these models were trained and how those texts were tokenized. There is speculation that the number of individual word tokens or word parts in the resulting texts was kept minimal, which introduces further bias when these tools are used to classify model tokens.

\paragraph{Language bias:} Many of these tools exhibit significant bias toward certain languages. For example, some logographic languages (such as Chinese) might be incorrectly identified when processing Cyrillic text. This bias likely stems from the training dataset and the unbalanced distribution of texts for different languages within that dataset.

\paragraph{Issues with less commonly used languages:} These tools often struggle with classifying tokens from less widely represented languages. This problem is also tied to the dataset, which likely contains very few texts in these languages, leading to biased classification of tokens belonging to these languages.

\subsubsection{Token Classification Issues}
Many of the current tools are not designed to handle token classification well, as tokens are parts of texts, and sometimes even parts of words.

\paragraph{Multi-language tokens:} Some tokens can belong to multiple languages, such as words that exist in several natural languages, as well as word fragments. Current tools struggle to assign multiple languages to a single token, since there is no convenient mechanism for ranking the likelihood of multiple languages being associated with one token.

\subsubsection{Missing Mapping Between Language Codes and Characters}
There is no standardized mapping of language codes to the characters used in each language, which means there is no cross-checking of classified languages against the actual characters present in the token. This makes it difficult to ensure accurate classification.

Certain languages, especially those without conventional delimiters (like spaces), pose unique challenges to classification and tokenization. For instance, languages like Chinese or Japanese, which use characters without spaces between words, make it difficult to apply standard tokenization and classification methods, requiring more specialized approaches.

These challenges highlight the limitations of current natural language classification tools, particularly when applied to token-based analysis or to languages that are less represented in common datasets. Addressing these issues will require both improvements in dataset diversity and more sophisticated methods for handling short or ambiguous tokens.

\section{Conclusion}

Based on our comprehensive analysis of tokenizers across various language models, we can draw several key conclusions that highlight the importance of tokenization in natural language processing and suggest directions for future research and development:

1. Tokenizer Diversity and Standardization: Our study revealed significant diversity among tokenizers, with 13 distinct tokenizers identified across 58 models. This diversity underscores the lack of standardization in tokenization approaches, which can impact model performance and cross-model compatibility.

2. Language Representation Bias: We observed a clear bias towards Latin script and major languages in most tokenizers. This bias potentially limits the effectiveness of models for less-represented languages and scripts, highlighting the need for more inclusive tokenization strategies.

3. Tokenizer as a Dataset Proxy: Our analysis suggests that tokenizers can serve as a latent representation of the training dataset. This insight opens up new possibilities for dataset analysis and potential reverse-engineering of training data characteristics.

4. Quality Control Tool: The development of Qtok, our unified tokenizer, provides a novel benchmark for evaluating tokenizer quality. This tool can significantly contribute to the standardization and improvement of tokenization practices across the field.

5. Tokenizer Improvement Potential: Our research indicates that there is considerable room for improvement in tokenizer design, particularly in expanding vocabulary size and improving representation for non-major languages.

6. Need for Specialized Tools: We identified a critical gap in tools for analyzing and classifying tokens, especially for minority languages and short text segments. Developing such tools could greatly enhance our ability to create more effective and inclusive tokenizers.

7. Transparency and Accessibility: The lack of access to proprietary tokenizers and training datasets hinders comprehensive analysis and improvement of tokenization techniques. Greater transparency in this area could accelerate progress in the field.

In conclusion, our study demonstrates that tokenization is a crucial, yet often overlooked, component in the development of language models. Improving tokenizer design and evaluation could lead to significant advancements in model performance, cost-efficiency, and language inclusivity. Future research should focus on developing standardized metrics for tokenizer quality, creating more inclusive tokenization strategies, and exploring the relationship between tokenizers and model performance across diverse languages and tasks. By addressing these challenges, we can pave the way for more effective and equitable natural language processing technologies.

\section{Availability}

We will provide our analysis tool and comprehensive result tables to ensure reproducibility and to support further research in this area.
Link to github, to pip, page on HF with table

In addition to GitHub, we have a collaborative notebook that enables full reproduction of the entire analysis, including the generation of all tables and figures. This notebook is available on GitHub.

Qtok availability: https://github.com/nup-csai/Qtok/

The reproducible code for all images and tables is available in our repository as a Jupyter notebook.

\section*{Acknowledgments}

We thank Alexander Avdiushenko for reading the manuscript and providing valuable suggestions and corrections, which significantly improved the quality of this work.

Thank you GPT4o and Claude 3.5 Sonnet for language editing.

\bibliographystyle{unsrt}

\end{document}